\documentclass[10pt,twocolumn,letterpaper]{article}

\usepackage[pagenumbers]{cvpr} 

\usepackage{graphicx}
\usepackage{amsmath}
\usepackage{amssymb}
\usepackage{booktabs}
\usepackage{booktabs}
\usepackage{amsfonts,amssymb}
\usepackage{ragged2e} 
\usepackage{booktabs,makecell, multirow, tabularx}
\usepackage[table]{xcolor}
\usepackage{colortbl}
\usepackage{array}

\usepackage[table]{xcolor}
\usepackage{colortbl}
\usepackage{xcolor}
\definecolor{myyellow}{rgb}{1.0, 0.49, 0.0}

\usepackage[pagebackref,breaklinks,colorlinks]{hyperref}

\usepackage[capitalize]{cleveref}
\crefname{section}{Sec.}{Secs.}
\Crefname{section}{Section}{Sections}
\Crefname{table}{Table}{Tables}
\crefname{table}{Tab.}{Tabs.}

\def\cvprPaperID{*****} 
\def\confName{CVPR}
\def\confYear{2025}

\begin{document}

\title{ID-NeRF: Indirect Diffusion-guided Neural Radiance Fields for \\ Generalizable View Synthesis}

\author{Yaokun Li,\quad   Chao Gou,\quad   Guang Tan\\
Shenzhen Campus of Sun Yat-sen University\\
{\tt\small liyk58@mail2.sysu.edu.cn, \{gouchao,tanguang\}@mail.sysu.edu.cn}
\\
}
\maketitle

\begin{abstract}
   Implicit neural representations, represented by Neural Radiance Fields (NeRF), have dominated research in 3D computer vision by virtue of high-quality visual results and data-driven benefits. However, their realistic applications are hindered by the need for dense inputs and per-scene optimization. To solve this problem, previous methods implement generalizable NeRFs by extracting local features from sparse inputs as conditions for NeRF decoder. However, although this way can allow feed-forward reconstruction, they suffer from the inherent drawback of yielding sub-optimal results caused by erroneous reprojected features. In this paper, we focus on this problem and aim to address it by introducing pre-trained generative priors to enable high-quality generalizable novel view synthesis. Specifically, we propose a novel Indirect Diffusion-guided NeRF framework, termed ID-NeRF, which leverages pre-trained diffusion priors as a guide for the reprojected features created by previous paradigm. Notably, to enable 3D-consistent predictions, the proposed ID-NeRF discards the way of direct supervision commonly used in prior 3D generative models, and instead adopting a novel indirect prior injection strategy. This strategy is implemented by distilling pre-trained knowledge into an imaginative latent space via score-based distillation, and an attention-based refinement module is then proposed to leverage the embedded priors to improve reprojected features extracted from sparse inputs. We conduct extensive experiments on multiple datasets to evaluate our method, and the results demonstrate the effectiveness of our method in synthesising novel views in a generalizable manner especially in sparse settings.
\end{abstract}
\begin{figure}[t]
	\centering
	\includegraphics[scale=0.85]{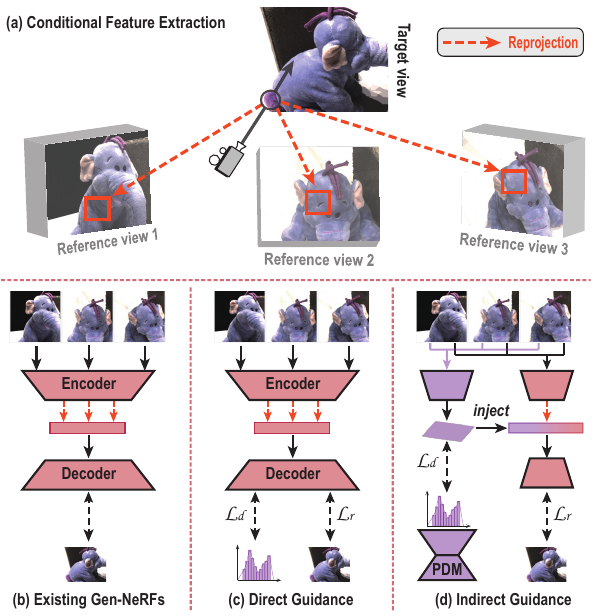}
	\caption{(a) A scenario illustrating the reprojection principle of Generalizable NeRFs; (b) The inference process of existing Gen-NeRFs; (c) Gen-NeRF under direct guidance using the modeled distribution; (d) Our model that uses indirect guidance. In this model, the reprojected features are refined using a diffusion-guided latent space (purple patch). $\mathcal{L}_s$ and $\mathcal{L}_r$ are score-based distillation loss and reconstruction loss, respectively.}
	\label{fig:graphical_abstract}
\end{figure}

\section{Introduction}\label{sec:intro}
With the development of AI technology, recent years have witnessed an increasing focus on implicit neural representations~\cite{deepsdf,nerf,occupancy} for 3D modelling in many classic 3D vision tasks, including but not limited to virtual try-on~\cite{virtual-try-on1,virtual-try-on2,virtual-try-on3}, scene understanding~\cite{su1,su2}, and avatar reconstructions~\cite{avatar1,avatar3}. This advancement is particularly noteworthy due to the substantial advantages these representations offer in terms of spatial resolution and representational capacity compared to traditional explicit representations. However, despite the advantages, these implicit representations, exemplified by NeRF, are constrained by the need for per-scene optimization with a considerable number of posed images, which is often impractical for real-world applications.

A line of research \cite{pixelnerf,ibrnet,NEO360,mvsnerf,nerfusion,lirf,dbarf,Neuralray, geonerf,GPNR} attempts to address the issue by implementing generalizable NeRF (Gen-NeRF) from sparse inputs. They achieve generalization by introducing scene-specific conditional features for NeRF, and the process of acquiring features can be illustrated in Fig.~\ref{fig:graphical_abstract}(a). For the feed-forward inference of the target view, 3D points from each target ray will be reprojected to the reference views to extract corresponding local features, which are then fused and used as conditions for the NeRF decoder. While this manner allows feed-forward inference by extracting geometric conditional features from sparse inputs, it has inherent limitations~\cite{Lin_2023_WACV,sparsefusion}. For example, when reconstructing the elephant's tail region in Fig.~\ref{fig:graphical_abstract}(a), the geometric reprojection of previous Gen-NeRFs will extract erroneous visual features, thus leading to a poor reconstruction of this region. Some previous methods~\cite{Neuralray,MatchNeRF} pointed out this problem and improved it by predicting the visibility of these reprojected features~\cite{Neuralray} or calculating the similarity information between them~\cite{MatchNeRF}. They work in most of the cases, but are still ineffective for reconstruction in unobserved regions of input views, because they can only utilize visual cues of input views, while some visual cues of the target view may be completely unobserved by input views. To solve this problem in essence, it is necessary to equip the generalizable NeRFs with the ability to generate new possible visual cues.

From this perspective, in this paper, we aim to inject generative capabilities into the current Gen-NeRFs to address the above challenges encountered with sparse input. We can draw inspiration from prior efforts \cite{NeRF-VAE,DiffusioNeRF,FeatureNerf} that use distributions generated by pre-trained generative models to directly supervise NeRF renderings via score-based distillation, as depicted in Fig. \ref{fig:graphical_abstract} (c). These methods can contribute to improving the quality of reconstruction, as the distribution generated by pre-trained generative models contains potential visual cues for unobserved regions. However, this manner may still encounter difficulties in producing photo-realistic renderings and suffer from the model confusion problem~\cite{multi-face-janus1,multi-face-janus2}. This limitation arises from the fact that views sampled from the generated distribution are often 3D inconsistency, and using them as the ground truth can lead to model confusion problems and favor smoothing predictions~\cite{northcutt2017learning, confident_learning, sparsefusion}.

To this end, we propose an novel indirect diffusion-guided generalizable NeRF framework to address the above issues. Instead of relying on direct supervision of pretrained diffusion models (PDMs), the main idea of our ID-NeRF, as illustrated in Fig. \ref{fig:graphical_abstract}(d), is to indirectly utilize the prior from PDMs through a distilled, imaginative latent space. This approach not only avoids model confusion but also empowers Gen-NeRFs with generative capabilities. Specifically, we first infer a latent code using a latent-space inference module. Subsequently, this latent code undergoes score-based distillation within a frozen PDM~\cite{stable-diffusion}, aimed at distilling knowledge from the PDM to this latent code, which may not be present in the input visual cues. After that, an attention-based refinement module is proposed for improving the reprojected features with this latent code, which are then fed into the implicit MLP as conditions for prediction and rendering. By doing so, our method avoids blurry rendering caused by inconsistent supervised signals and is capable of producing photorealistic predictions from sparse inputs. We have conducted extensive experiments to validate our design, and the results demonstrate that our method achieves superior results compared with state-of-the-art (SOTA) methods. In summary, our primary contributions are as follows:

\begin{itemize}
	\item We propose ID-NeRF, which improves the long-standing suboptimal issue of erroneous reprojection in sparse settings observed in previous generalizable NeRFs by introducing the pre-trained generative priors. 
	\item We introduce an indirect prior injection strategy, by implementing score-based distillation on an intermediate latent space instead of on NeRF renderings, which improves the rendering quality of generalizable novel view synthesis while avoiding the model confusion problem.
	\item We conduct qualitative and quantitative experiments on multiple datasets, and the results demonstrate the effectiveness of our method.
\end{itemize}

\section{Related Work}

\subsection{Generalizable NeRF}

NeRF~\cite{nerf} in its basic form is an MLP-based mapping function for predicting rendering attributes like color and density with scene-agnostic inputs of coordinates and orientations. Therefore, it needs to be retrained when rendering a new scene, since the mapping function cannot produce different content for the same inputs. To address this issue, subsequent works \cite{pixelnerf,ibrnet,mvsnerf,Neuralray, geonerf,GPNR,nerfusion,lirf,dbarf} have attempted to make NeRF generalizable by adding scene-dependent discriminative conditions as additional inputs.

Pixel-NeRF \cite{pixelnerf} is among the pioneering efforts in this direction. It constructs a feature volume and extracts image features from it to serve as conditions for NeRF. IBRNet \cite{ibrnet} proposes a weighted MLP network to process the local features extracted from nearby views, followed by a ray transformer and an MLP to predict density and color. MVSNeRF \cite{mvsnerf} utilizes a 3D CNN network to process a constructed 3D cost volume, and then uses a conditional MLP to predict color and density. A recent approach, DBARF \cite{dbarf}, trains a cost feature map in a self-supervised manner and generalizes by projecting and interpolating local features, similar to IBRNet.

These methods extract local features as discriminative conditions for each scene through reprojection. This implies that they rely on visual cues present in the reference views. As a result, when the input views are sparse or differ significantly from the target view, the reprojected visual cues for unseen regions will likely be erroneous. Some works try to address the uncertainty by incorporating additional geometric constraints. For instance, NeuRay~\cite{Neuralray} constructs a visibility feature vector for each input view to predict the visibility of each 3D point to enable occlusion-aware feature aggregation, and MatchNeRF~\cite{MatchNeRF} computes similarity information for the reprojected local features and concatenates them for enhancement. Nevertheless, these filter-based approaches does not fundamentally improve the visual cues, since it cannot generate new and reasonable content for unobserved regions beyond what is already present in the reference views.

\subsection{NeRF with Generative Models}

Motivated by the recent development of generative modelling in the field of 2D vision, there is a body of work that treats the task of novel view synthesis as a generative task. They typically combine NeRF with generative models with the aim of giving NeRF the ability to imagine unobserved visual cues. Previous work attempts to integrate GANs~\cite{Giraffe,Graf,pi-gan} or VAEs~\cite{NeRF-VAE} with NeRFs, but they suffer from either insufficient controllability or poor representational capabilities.

Recently, with diffusion models~\cite{DDPM,DDIM}, renowned for more controllable and high quality generation, as the new paradigm for generative models, there is an increasing number of works focusing on the combination of diffusion with NeRF. DreamFusion~\cite{dreamfusion} is one of their pioneering works, which introduces the Score Distillation Sampling (SDS) loss to bridge the gap between using pre-trained 2D diffusion models~\cite{Imagen,stable-diffusion} to guide NeRF training. Inspired by it, subsequent generative NeRF works~\cite{dreambooth3d,dream3d,magic3d,compositional,sjc,Zero-1-to-3,One-2-3-4} can achieve considerable 3D reconstruction conditional on text or a single image. However, their direct supervision manner faces an unavoidable 3D inconsistency problem. That is, they sample views from the generated distribution as the ground truth, which may correspond to inconsistent 3D objects, such as two elephants with purple and red tails. These inconsistent supervision signals can confuse the model and result in ambiguous predictions. In addition, there is some work that does not rely on pre-training diffusion prior but rather re-trains a diffusion model. For instance, DiffusioNeRF~\cite{DiffusioNeRF} generates a scene prior using a denoising diffusion model to supervise NeRF's color and density predictions. SparseFusion~\cite{sparsefusion} first extracts objects using pre-defined masks and then jointly trains a NeRF and diffusion model. However, they still belong to the direct supervision paradigm and require expensive per-scene optimization or time-consuming mask annotations, which hinders their application in real-world scenarios.

In this paper, we draw inspiration from the aforementioned works and aim to improve the suboptimal issue caused by erroneous reprojections in current generalizable NeRFs from a generative perspective. To avoid the challenges faced by prior works in introducing generative priors, we propose an indirect strategy to inject prior from pre-trained diffusion models without the need for additional mask labels or retraining of diffusion models. In previous work, LatentNeRF~\cite{latent-nerf} performs SDS loss in a latent space similarly to ours, but they are fundamentally different from ours by decoding latent directly into novel views and requiring text as input as well as no generalization capability.


\section{Preliminaries}
\label{pre} 

\textbf{Generalizable NeRF.} Vanilla NeRF \cite{nerf} maps the coordinate $x$ and ray direction $d$ of a 3D point to its density $\sigma$ and color $c$ using a two-stage MLP network $\mathcal{M_\theta}:(x,d)\to(\sigma,c)$. The below volume rendering is then used to compute the color of each emitted ray by integrating over the N sampling points within its near and far bounds $[t_n,t_f]$:
\begin{equation}\label{volume-render}
	C({\rm r}) = \sum_{i=1}^{N}T_i(1-{\rm exp}(-\sigma_i \delta_i))c_i
\end{equation}
where $ T_i={\rm exp}(-\sum_{j=1}^{i-1}\sigma_j \delta_j )$ is the volume transmittance and $\delta_i = t_{i+1}-t_i$ is the interval distance. The reconstruction loss is performed on $C({\rm r})$ for optimization.

For different scenes, the inputs $x, d$ are the same for the above manner and hence the vanilla NeRF cannot be generalized. Therefore the current Gen-NeRFs inputs scene-specific geometric information into $\mathcal{M_\theta}$ as discriminative condition. Given a scene with N input views $\mathcal{I}=\{I_i\}_{i=1}^{N}$ and corresponding camera intrinsic and extrinsic parameters $\mathcal{P}=\{P_i\}_{i=1}^{N}$, for predicting the color and density of a 3D point of a novel view $I$, Gen-NeRFs will first project it to $N$ input views to extract scene geometric features:
\begin{equation}
	z = \mathcal G(\{p(x,P_i),b(I_i)\}_{i=1}^{N})
\end{equation}
where $\mathcal G(\cdot)$ is the aggregation function, different for different methods. $b(\cdot)$ is the backbone. $p(\cdot)$ is the projection operation, which utilizes the camera parameter $P_i$ to project the 3D point $x$ into 2D coordinates on each reference view plane and then extracts the corresponding feature values, as described in more detail in~\cite{pixelnerf,contranerf}. The geometric condition $z$ is then entered into $\mathcal{M_\theta}$ along with the coordinate $x$ and direction $d$ to predict the rendering factor.
\begin{equation}
	\sigma, c= {\mathcal M}_\theta(x,d,z)
\end{equation}	

\begin{figure*}[t]
	\centering
	\includegraphics[scale=0.52]{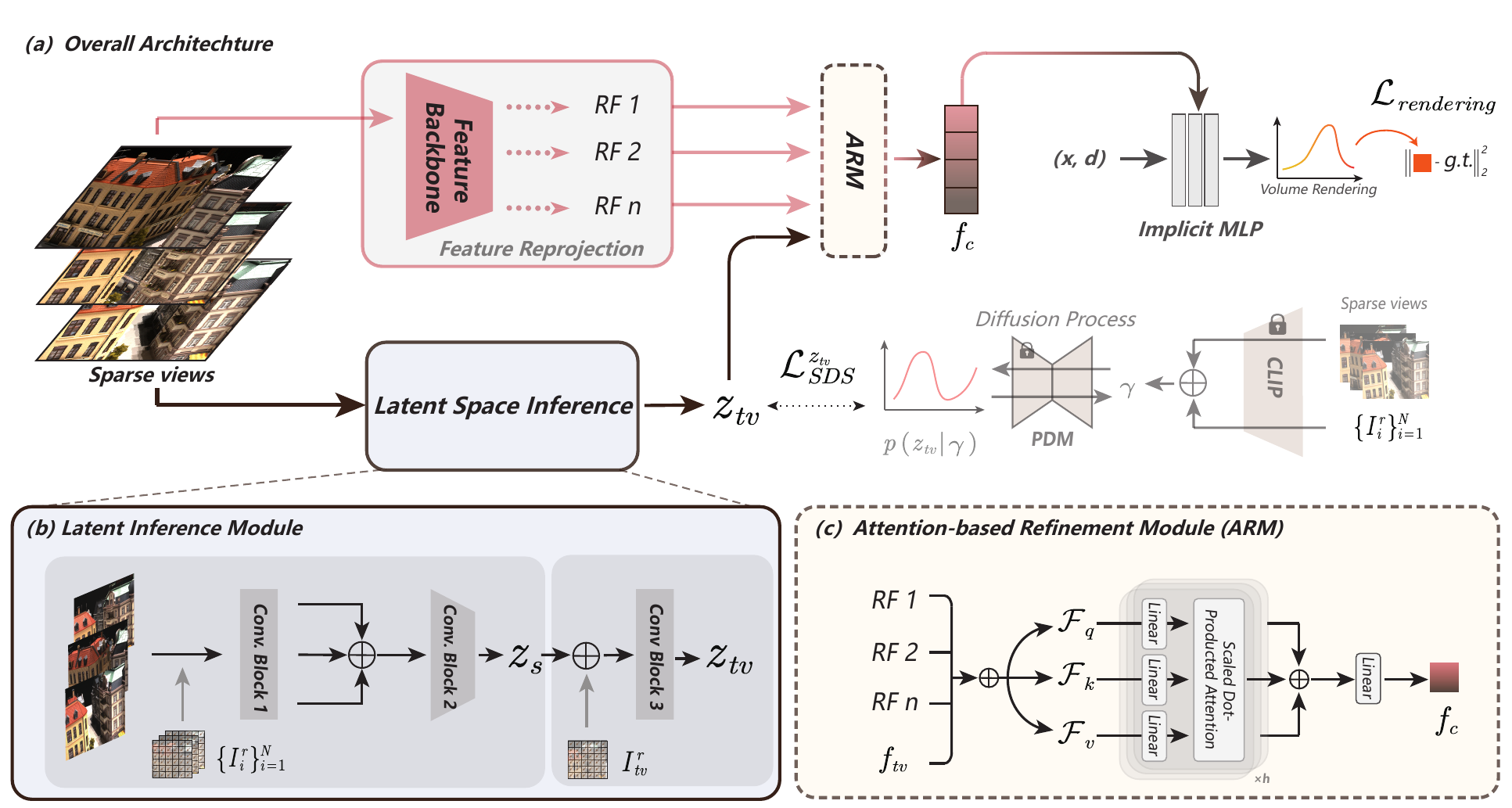}
	\caption{Overview of our ID-NeRF. Given sparse views, there are two workflows to process them. The first (red) utilizes geometric reprojection to obtain reprojected features (RF). The other one (black) uses an inference module to predict the latent space $z_{tv}$, which is performed score-based distillation with the PDM-predicted distribution $p(z_{tv}|\gamma)$. Then, these features are fed together into the ARM to obtain the refined conditional feature $f_{c}$. $\{I^r_i\}_{i=1}^N$ and $I^r_{tv}$ are ray images used to enhance the pose information.}
	\label{fig:overview}
\end{figure*}

\textbf{Score-based Distillation.}  Denoising diffusion probabilistic models  \cite{DDIM,DDPM} (DDPM), or score-based generative models \cite{score-based-1,score-based-2} have recently shown remarkable success in content generation \cite{DALLE-2,Imagen,stable-diffusion}. In the field of text-to-3D, in order to utilize the knowledge of pre-trained DDPMs to bootstrap 3D generation, DreamFusion \cite{dreamfusion} pioneered the Score Distillation Sampling (SDS) loss, on which most of the subsequent works \cite{magic3d,dream3d,dreambooth3d,3DFuse} have been implemented. For a NeRF-rendered image $I$, the SDS optimization first performs a diffusion process on it, i.e., noising it to a standard Gaussian distribution in $t$ time steps. 

\begin{equation}
	I_t=\sqrt{\bar{\alpha}_t} I+\sqrt{1-\bar{\alpha}_t} \epsilon 
\end{equation}

In the reverse (generative) process, the pre-trained DDPM generates $t$ noises under the text condition $\gamma$ to continuously denoise the standard Gaussian distribution. Then, the score-based distillation gradient of the rendered image can be obtained as follows, where $\epsilon_\phi$ is the pre-trained denoiser.
\begin{equation}
	\mathcal{L}_{SDS} := {\mathbb E_{I,t,\gamma,\epsilon \sim {\mathcal N}(0,1)}}[\left \| \epsilon_\phi(I_{t},t,\gamma)-\epsilon  \right \|_2^2 ]
\end{equation}

\section{METHODOLOGY}

We propose ID-NeRF, a generalizable NeRF framework under the indirect guidance of a pre-trained diffusion model~\cite{stable-dreamfusion}. An overview of our approach is illustrated in Fig. \ref{fig:overview}, and as seen, there are two workflows in our method for processing multi-view images, which yields the reprojected features and latent code respectively. After this, an attention based module uses the latent code to improve these reprojected features to produce refined condition features, which are further fed into the NeRF decoder to predict the corresponding colors and densities. In this section, we will unfold our approach step by step.

\subsection{Scene-specific Geometric Information Extraction}

Given $N$ input views $\mathcal{I}=\{I_i\}_{i=1}^{N}$, our method first extracts reprojected features like typical generalized NeRFs. In the case of sparse setting, the geometric matching prior has been demonstrated to be important in many works~\cite{LoFTR,mvsnerf,MatchNeRF}. Therefore, we take GMFlow's Transformer~\cite{gmflow} as our backbone to obtain geometric-aware features $\{F_i\}_{i=1}^N$. To render a target view $I_{tv}$, we sample n rays on it and m 3d points on each ray. Then, for a 3D sample point, we project it onto feature planes $\{F_i\}_{i=1}^N$ by camera parameters to extract the corresponding local features $\{f_i\}_{i=1}^N$. In addition, we also extract the local colors $\{c_i\}_{i=1}^N$ and compute the mutual similarity $\{s_i\}_{i=1}^N$ between $\{f_i\}_{i=1}^N$ as additional geometric information as in MatchNeRF~\cite{MatchNeRF}. Finally, these information are concatenated together along the viewpoint dimension to obtain the scene-specific reprojected feature $\{p_i\}_{i=1}^N$ for this 3D sample point.

\subsection{Latent Space Inference Assisted by Distilling PDM}\label{3C}

Another stream of data processing represents the core of our approach, which infers the latent space with the aid of distilling a PDM~\cite{stable-diffusion}. Firstly, a latent inference module $\mathcal{E}$ is used for inferring a low-dimensional latent space $z$ from input visual cues. Then, we condition the pre-trained diffusion model with sparse views $\mathcal{I}$ to perform score-based distillation on the inferred $z$. Note that the pre-trained diffusion model's branch is only used during training, i.e., at inference time we directly use the latent inference model to predict $z$ and perform the subsequent steps. Such a process is a somewhat like distilling the generative power of the PDM into the latent inference model during training.

Specifically, the latent inference module $\mathcal{E}$ contains two stages that guarantee latent space inference from the scene-level to the view-level. The first stage is similar to an auto-encoder with the aim at inferring a scene-level latent containing the entire scene information. It involves two convolution blocks, each containing three convolution layers. To achieve better prediction, we add camera pose information to each view before feeding it to the CNN blocks. Concretely, for each input view $I_i$, we create a ray image $I^r_i = \mathcal{R}_d - \mathcal{R}_o$ to concatenate, where \(\mathcal{R}_d, \mathcal{R}_o \in \mathbb{R}^{3 \times H \times W}\) are the direction and origin of the ray at each pixel calculated by the camera parameters, respectively. These enhanced images are encoded by a 3-layer weight-sharing convolution block to get geometric features, respectively. Then, we concatenate these features along the feature dimension and use another 3-layer convolution block to get the scene-level latent $z_s$. 
\begin{table}[h]
	\caption{Parameters of convolution neural networks in the latent inference module.}
	\label{cnn}
	\centering
	\renewcommand{\arraystretch}{0.92}
	\small
	\setlength{\tabcolsep}{1,0mm}{\begin{tabular}{cccccc}
			\toprule
			\multirow{2}{*}{Blocks} & Layer  & Channels & Kernel & \multirow{2}{*}{Stride} & \multirow{2}{*}{Padding}\\
			& number  & (in, out) & size &  & \\
			\midrule
			\multirow{3}{*}{Conv. Block 1} &  1.1 & (6, 16)  & 3x3  & 2 & 1 \\
			&  1.2 & (16, 32)  & 3x3  & 1 & 1 \\
			&  1.3 & (32, 32)  & 3x3  & 1 & 1 \\
			
			\midrule
			\multirow{3}{*}{Conv. Block 2} &  2.1 & (96, 64)  & 3x3  & 2 & 1 \\
			&  2.2 & (64, 64)  & 3x3  & 2 & 1 \\
			& 2.3 & (64, 32)  & 3x3  & 1 & 1 \\
			
			\midrule
			\multirow{6}{*}{Conv. Block 3} &  3.1 & (35, 32)  & 3x3  & 2 & 1 \\
			&  3.2 & (32, 32)  & 3x3  & 2 & 1 \\
			&  3.3 & (32, 16)  & 3x3  & 2 & 1 \\
			&  3.4 & (16, 16)  & 3x3  & 1 & 1 \\
			\bottomrule[1pt]
	\end{tabular}}
\end{table}	

It should be noted that although this inferred latent $z_s$ can ideally contain information about the entire scene, it may be somewhat noisy for synthesizing a specific view. Therefore, in order to provide fine-grained guidance at the target view level, we further introduce another convolution block in the second stage for inferring a view-level latent. Before this, we produce a ray image $I^r_{tv}$ in the same way and concatenate it to reshaped latent ${z}'_s$ of the same size. Thereafter, we encode them using a 4-layer convolution block to get $z'_{tv}$ and finally a fully-connected layer is applied on the feature dimension to reduce the dimension from 16 to 4 to get the final view-level latent $z_{tv} \in \mathbb{R}^{4 \times h \times w}$, where $h=w=64$. The parameters of the above mentioned convolution neural networks can be found in Table~\ref{cnn}.

After obtaining latent $z_{tv}$, we will perform the following SDS loss for it during training. As described in Sec.~\ref{pre}, in the forward process, we gradually add noise to latent $z_{tv}$ in $t$ steps until it becomes a standard Gaussian distribution. In the backward process, we use a pre-trained stable diffusion model conditioned on $\gamma$ to generate t noises, which are used to recover latent $z_{tv}$ from a standard Gaussian distribution. To get the condition, we first use a pre-trained CLIP encoder~\cite{CLIP} to separately encode the sparse views $\{I_i\}_{i=1}^N$ as well as their corresponding ray image $\{I^r_i\}_{i=1}^N$, and then concatenate their embeddings as $\gamma$. Finally, $z_{tv}$ is optimized by Eq.~\ref{diffusion}, where $\epsilon_\phi$ is the PDM's denoiser and $z_{{tv}_t}$ is the noisy latent code at step $t$. Note that the parameters of the CLIP and stable diffusion model used here are frozen, and therefore, they are not involved in gradient updating during training. Additionally, the above diffusion guidance process is not performed during testing.

\begin{equation}\label{diffusion}
	\mathcal{L}^{z_{tv}}_{SDS}= {\mathbb E_{z_{tv},t,\gamma,\epsilon \sim {\mathcal N}(0,1)}}[\left \| \epsilon_\phi(z_{{tv}_t},t,\gamma)-\epsilon  \right \|_2^2 ]
\end{equation}	

\subsection{Attention-based Refinement for Reprojected Visual Cues}
After implementing the above process, we now get the reprojected visual cues $\{p_i\}_{i=1}^N$ and a view-level latent $z_{tv}$. Here we propose an Attention-based Refinement Module (ARM) for integrating them, with the aim of using latent $z_{tv}$ to improve these reprojected features. Specifically, we first use two fully-connected layers to compress $z_{tv}$ into a vector $f_{tv}$, which has the same dimension as $p_i$, and then concatenate it with $\{p_i\}_{i=1}^N$ as a vector sequence $x$. Then, the sequence $x\in\mathbb{R}^{(N+1)\times d}$ is input into a $m$-layer multi-head self-attention (MSA) module \cite{transformer} to adaptively generate weights for fusion. The MSA can be formulated as follows, where $W_q,W_k,W_v \in \mathbb{R}^{d\times d_p}$ and $W_m \in \mathbb{R}^{hd_p\times d}$ are the learnable weight matrices and h is the number of heads.
\begin{equation}
	SA(x) = softmax(\frac{xW_{q}(xW_{k})^{\top}}{\sqrt{d_{k} }} )xW_{q}
\end{equation}	
\begin{equation}
	MSA(x) = cat(SA_{1}(x),\dots,SA_{h}(x))xW_{m}
\end{equation}		

\begin{table*}[h]
	\caption{Quantitative results on different datasets for three input views.}
	\label{main results 1}
	\centering
	\renewcommand{\arraystretch}{1.3}
	\small
	\setlength{\tabcolsep}{4,3mm}{\begin{tabular}{ccccccc}
			\toprule
			\multirow{2}{*}{\textbf{Methods}}&\multicolumn{3}{c}{\textbf{DTU}}&\multicolumn{3}{c}{\textbf{RFF}}\\ 
			\cmidrule(r){2-4} \cmidrule(r){5-7} 
			&\textbf{PSNR}$\uparrow$&\textbf{SSIM}$\uparrow$&\textbf{LPIPS}$\downarrow$&\textbf{PSNR}$\uparrow$&\textbf{SSIM}$\uparrow$&\textbf{LPIPS}$\downarrow$\\ 
			\midrule			
			
			PixelNeRF \cite{pixelnerf} & {19.31}  & {0.789}  & {0.382} & 11.24 & 0.486 & 0.671 \\
			
			IBRNet \cite{ibrnet} & {26.04}  & {0.917}  & {0.190} & 21.79 & 0.786 & 0.279 \\
			
			GNT \cite{GNT} & {26.39}  & {0.923}  & \cellcolor{myyellow!17}{0.156} &  \cellcolor{myyellow!17}{22.98} & {0.761} & \cellcolor{myyellow!17}{0.221} \\		
			
			MVSNeRF~\cite{mvsnerf} & \cellcolor{yellow!17}{26.63}  & \cellcolor{yellow!17}{0.931}  & {0.168}& 21.93 &\cellcolor{yellow!17}{0.795}& {0.252}\\
			
			MatchNeRF~\cite{MatchNeRF} & \cellcolor{myyellow!17}{26.91}  & \cellcolor{myyellow!17}{0.934}  & \cellcolor{yellow!17}{0.159}& \cellcolor{yellow!17}{22.43}& \cellcolor{myyellow!17}{0.805} &\cellcolor{yellow!17}{0.244} \\
			\midrule
			ID-NeRF(Ours) & \cellcolor{red!18}{27.19}  & \cellcolor{red!18}{{0.937}}  & \cellcolor{red!18}{0.150} & \cellcolor{red!18}{23.07} & \cellcolor{red!17}{0.812} &\cellcolor{red!17}{0.212}  \\
			
			\bottomrule[1pt]
	\end{tabular}}
\end{table*}	

\subsection{Optimization}
\textbf{Volume Rendering.} The purpose of ID-NeRF is to improve the reprojected features in typical Gen-NeRFs with an imaginative latent. Therefore, in order to demonstrate the effectiveness of our method, we do not make improvements in the NeRF decoder module and use the same MLP network as~\cite{MatchNeRF}. Then, the colors and densities of all sampled points on a ray are rendered by Eq. \ref{volume-render} to get the corresponding projected colors $C(r)$.

\textbf{Training objective.} After rendering, the mean square error loss is implemented on each sampled ray, where $\tilde{C}(r)$ is the ground truth color.
\begin{equation}\label{geo-loss}
	\mathcal{L}_{rendering}=\sum_{r\in \mathcal{R}}^{} \left \| C(r) - \tilde{C}(r) \right \|_2^2
\end{equation}

Thus, the total loss of our method is as follows, where $\lambda_s$ and $\lambda_r$ are the weight values.

\begin{equation}
	\mathcal{L}_{total} = \lambda_s\mathcal{L}^{z_{tv}}_{SDS}+\lambda_r\mathcal{L}_{rendering}
\end{equation}

\begin{figure*}[t]
	\centering
	\includegraphics[scale=0.59]{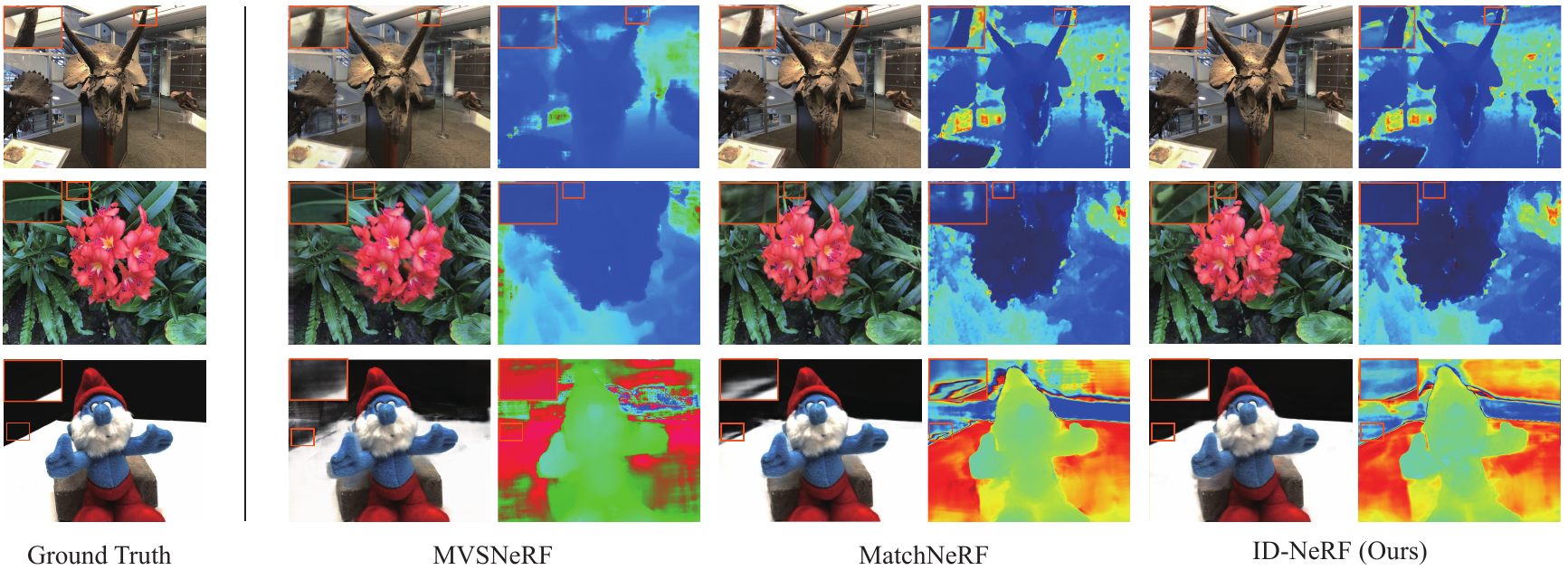}
	\caption{Qualitative comparison of rendering results. We present the rendered RGB images and depth maps of our ID-NeRF as well as representative MVSNeRF \cite{mvsnerf} and MatchNeRF \cite{MatchNeRF}, with each result zoomed in on details.}
	\label{fig:visualization_main_results}
\end{figure*} 

\section{Evaluation}
\subsection{Experimental Settings}

\textbf{Datasets and Metrics.} In comparison with other methods, we follow the experimental protocol of MVSNeRF~\cite{mvsnerf}, i.e., training on 88 scenes and then testing on another 16 scenes on the DTU \cite{DTU} dataset. In addition, we perform tests on another real-world scene-level dataset, Real Forward-Facing (RFF)~\cite{nerf}. In our experiments, we take two different sparse settings of three and two input views for training. To measure the effectiveness of the models, we take PSNR, SSIM \cite{SSIM} and LPIPS \cite{LPIPS} as metrics as in previous arts. 

\textbf{Baselines.} To demonstrate the effectiveness of our method, we compare ID-NeRF with the current SOTAs of reprojection-based Gen-NeRFs, including PixelNeRF \cite{pixelnerf}, IBRNet \cite{ibrnet}, MVSNeRF \cite{mvsnerf}, MatchNeRF \cite{MatchNeRF}, and GNT~\cite{GNT}. To verify the effectiveness of our method in the task of generalizable novel view synthesis with sparse inputs, we fist conduct tests in the manner commonly used in previous Gen-NeRFs~\cite{mvsnerf}, i.e., picking sparse views near the target view as input for reconstruction. However, while this setup has relatively few (3 or 2) input views, the relative poses between input views and the target view are small, so the degradation issue caused by erroneous reprojected features is still relatively minor. Therefore, to further validate our effectiveness, we also perform comparisons in more sparse scenarios, as detailed in Sec.~\ref{Results}.

\textbf{Implementation Details.}  
We implement our ID-NeRF on a single NVIDIA A100 GPU with 600k steps, sampling 1024 rays for training and 4096 rays for testing. In the MSA module, $m, d, dp$, and $h$ are 3, 8, 4, and 4, respectively. AdamW \cite{AdamW} algorithm and one cycle policy \cite{OneCycle} are adopted for optimization. We keep a learning rate of 5e-5 for GMFlow and NeRF decoder, and 1e-3 for the remaining modules. We implement SDS loss based on the codebase \cite{stable-dreamfusion} utilizing stable-diffusion (SD) with version v2-1. During training, we set $\lambda_s = \lambda_r = 1$.

\subsection{Experimental Results}\label{Results}
In this subsection, we perform two experimental setups comparing our method with baselines. We conduct comparisons first in the test setup commonly used in previous methods, and then in more sparse scenarios.

\textbf{Sparse setting 1.} In this setup, we follow the comparison method of MVSNeRF~\cite{mvsnerf}, inputting 3 or 2 nearby views separately for reconstruction. Firstly, for the experiment trained and tested with 3-input views, we report quantitative and qualitative results on the DTU and RFF datasets in Table \ref{main results 1} and Fig.~\ref{fig:visualization_main_results}, respectively. As shown in Table \ref{main results 1}, our method achieves the best results relative to baselines in all three metrics. As seen, PixelNeRF~\cite{pixelnerf} works badly in this setting, which is likely due to its reliance on a large number of input images. The follow-up IBRNet~\cite{ibrnet}, GNT~\cite{GNT}, and MVSNeRF~\cite{mvsnerf} improve on this, but they produce ambiguous predictions in certain regions due to their reliance on limited visual cues, leading to the sub-optimal performance in Table~\ref{main results 1}. For analysis in Fig. \ref{fig:visualization_main_results}, these methods often produce sub-optimal or incorrectly predicted colors and depths in regions near the image boundaries, whereas producing relatively better results at the central part of the image.

This can be explained by the process of changing camera poses. When the camera pose shifts from one view to another, the visual cues of the boundary in the original view are often not captured by the camera in the new view. Thus, for the boundary regions of the target view, their visual cues are also likely to be absent in the nearby input views. In other words, when reconstructing these areas, the features obtained through reprojection are likely to be erroneous, leading to the suboptimal results shown in Fig. \ref{fig:visualization_main_results}. Besides these methods, MatchNeRF~\cite{MatchNeRF} enhances the visual cues through geometric matching, which can improve the reconstruction quality in most areas, but as can be seen, the aforementioned sub-optimal reconstruction problem still occurs in the marginal areas. On this basis, we enhance the visual cues by injecting the powerful knowledge of PDM to be more sensitive to the boundary information, making it possible to generate more accurate depth maps while ensuring clear contours, as illustrated in Fig. \ref{fig:visualization_main_results}. In contrast, our approach injects a generative prior that improves these reprojected features by a latent space $z$ containing possible visual cues, which enables realistic reconstruction quality even at the boundaries.

In addition, we also compare our method with baselines in an experimental setup with 2-input views. In this experimental setup, we select GNT~\cite{GNT}, MVSNeRF~\cite{mvsnerf}, and MatchNeRF~\cite{MatchNeRF}, which have sub-optimal performance above, for comparison. The results are shown in Table~\ref{main results 2}, where our method still achieves SOTA performance, such results demonstrate the superiority of our method in generalizable novel view synthesis under sparse conditions
\begin{table}[t]
	\caption{Quantitative results on different datasets for two input views.}
	\label{main results 2}
	\centering
	\renewcommand{\arraystretch}{1.6}
	\small
	\setlength{\tabcolsep}{0,2mm}{\begin{tabular}{ccccccc}
			\toprule
			\multirow{2}{*}{\textbf{Methods}}&\multicolumn{3}{c}{\textbf{DTU}}&\multicolumn{3}{c}{\textbf{RFF}}\\ 
			\cmidrule(r){2-4} \cmidrule(r){5-7} 
			&\textbf{PSNR}$\uparrow$&\textbf{SSIM}$\uparrow$&\textbf{LPIPS}$\downarrow$&\textbf{PSNR}$\uparrow$&\textbf{SSIM}$\uparrow$&\textbf{LPIPS}$\downarrow$\\ 
			\midrule			
			
			GNT~\cite{GNT} & \cellcolor{yellow!17}{24.32}& {0.903}& {0.201}& \cellcolor{myyellow!17}{20.91}& {0.683}& {0.293} \\			
			
			MVSNeRF~\cite{mvsnerf} & 24.03 & \cellcolor{yellow!17}{0.914}& \cellcolor{yellow!17}{0.192}& 20.22 &\cellcolor{yellow!17}{0.763}& \cellcolor{yellow!17}{0.287}\\
			
			MatchNeRF~\cite{MatchNeRF} & \cellcolor{myyellow!17}{25.03} &\cellcolor{myyellow!17}{0.919} &\cellcolor{myyellow!17}{0.181}& \cellcolor{yellow!17}{20.59}& \cellcolor{myyellow!17}{0.775} &\cellcolor{myyellow!17}{0.276} \\
			
			ID-NeRF(Ours) & \cellcolor{red!17}{25.32} &\cellcolor{red!17}{0.928} & \cellcolor{red!17}{0.171} & \cellcolor{red!17}{20.95} & \cellcolor{red!17}{0.823} &\cellcolor{red!17}{0.194}  \\
			\bottomrule[1pt]
	\end{tabular}}
\end{table}	
\begin{table*}[ht]
	\caption{Comparison at different sparsity levels of input. \textbf{P.} means PSNR, \textbf{S.} means SSIM, \textbf{L.} means LPIPS. $\dagger$ denotes that the result is our reproduction based on the description or code of the paper.}
	\label{different sparse inputs}
	\centering
	\renewcommand{\arraystretch}{1.45}
	\small
	\setlength{\tabcolsep}{4,2 mm}{\begin{tabular}{c|ccc|ccc|ccc}
			\toprule
			\multirow{2}{*}{\textbf{Methods}}&\multicolumn{3}{c}{\textbf{Sparsity 1}}&\multicolumn{3}{c}{\textbf{Sparsity 2}}&\multicolumn{3}{c}{\textbf{Sparsity 3}}\\ 
			\cmidrule(r){2-4} \cmidrule(r){5-7}  \cmidrule(r){8-10}
			&\textbf{P.}$\uparrow$&\textbf{S.}$\uparrow$&\textbf{L.}$\downarrow$&\textbf{P.}$\uparrow$&\textbf{S.}$\uparrow$&\textbf{L.}$\downarrow$&\textbf{P.}$\uparrow$&\textbf{S.}$\uparrow$&\textbf{L.}$\downarrow$\\ 
			\midrule			
			
			PixelNeRF$\dagger$ \cite{pixelnerf} & {18.21}  & {0.739}  & {0.370} & 8.54 & 0.515 & 0.554 & 5.61 & 0.457 & 0.612 \\		
			
			IBRNet$\dagger$ \cite{ibrnet} & {25.62}  & {0.860}  & {0.183} & 14.82 & 0.654 & 0.347 & 11.91 & 0.639 & 0.403 \\
			
			GNT$\dagger$ \cite{GNT} & {25.94}  & \cellcolor{yellow!17}{0.891}  & \cellcolor{yellow!17}{0.147} &  {16.37} & {0.697} & \cellcolor{yellow!17}{0.298}  &  {13.49} & {0.633} & {0.350} \\		
			
			MVSNeRF$\dagger$~\cite{mvsnerf} & \cellcolor{yellow!17}{26.07}  & {0.887}  & {0.160}& \cellcolor{yellow!17}{16.89} &\cellcolor{yellow!17}{0.711}& {0.311}& \cellcolor{yellow!17}{14.07}& \cellcolor{yellow!17}{0.651}& \cellcolor{yellow!17}{0.339}\\
			
			MatchNeRF$\dagger$~\cite{MatchNeRF} & \cellcolor{myyellow!17}{26.23}  & \cellcolor{myyellow!17}{0.896}  & \cellcolor{myyellow!17}{0.143}& \cellcolor{myyellow!17}{17.18}& \cellcolor{myyellow!17}{0.725} &\cellcolor{myyellow!17}{0.278}& \cellcolor{myyellow!17}{14.42}& \cellcolor{myyellow!17}{0.674} &\cellcolor{myyellow!17}{0.322} \\
			
			ID-NeRF(Ours) & \cellcolor{red!18}{26.48}  & \cellcolor{red!18}{{0.897}}  & \cellcolor{red!18}{0.141} & \cellcolor{red!18}{17.98} & \cellcolor{red!17}{0.734} &\cellcolor{red!17}{0.270}& \cellcolor{red!18}{15.79} & \cellcolor{red!17}{0.688} &\cellcolor{red!17}{0.318}  \\
			\bottomrule[1pt]
	\end{tabular}}
\end{table*}	
\textbf{Sparse setting 2.} Further, it is important to note that the measurement standard for the above results takes the nearby views as inputs. This means that the input views are few but contain many regions of the target view. In other words, there are few unobserved regions in the above settings so the problem of incorrectly reprojected features has actually little impact.  

Therefore, to fully investigate the effectiveness of our method in improving erroneous reprojected features, we here conduct experiments in a more sparse input setting. Specifically, we still input three views, but unlike Setting 1, where nearby views are selected as inputs, we select views progressively farther from the target view for reconstruction. For simple and effective testing, we set up three different levels of sparsity in the 16 test scenes of the DTU dataset. In each sparsity level, we select a fixed view number (view 44) from these scenes as the target view, and the input view numbers for sparsity levels 1, 2, and 3 are (43, 33, 31), (26, 22, 10), and (22, 10, 3), respectively. These input views are progressively sparser relative to the target view, thereby increasing the impact of erroneous projections.

The results are reported in Table \ref{different sparse inputs}, from which we can see that as the inputs get sparser (from sparsity levels 1 to 3), our method results in a larger performance advantage over previous arts. For example, at sparsity 1, our method outperforms MatchNeRF by 0.25 in PSNR, and then by 0.80 and 1.37 at sparsity 2 and 3, respectively. This phenomenon suggests that the sparser the input, the better the relative performance of our method. Alternatively, our method is relatively less susceptible to erroneous reprojection at sparser inputs.

\subsection{Ablation Studies}
\textbf{Effect of latent space.} To evaluate the effectiveness of the latent space, here we perform ablation experiments on it. We set up three cases and report the results in Table~\ref{ablation: latent space}. $No\verb|-|latent$ indicates that only reprojected features are used as conditional features, $z_s$ indicates that scene-level latent is used to improve reprojected features, and $z_{tv}$ indicates that view-level latent is used to improve reprojected features. From the experimental results, it can be seen that as the latent space is introduced and its granularity is refined, the model's performance continuously improves, which proves the effectiveness of the introduction of the latent space.

\begin{table}[h]
	\caption{Ablation experiments of latent space on the DTU dataset under the three-input view setting.}
	\label{ablation: latent space}
	\centering
	\renewcommand{\arraystretch}{1.5}
	\small
	\setlength{\tabcolsep}{3,6mm}{\begin{tabular}{cccc}
			\toprule
			\textbf{Cases} & \textbf{PSNR$\uparrow$}  & \textbf{SSIM$\uparrow$} & \textbf{LPIPS$\downarrow$}\\ 
			\midrule
			
			$No\verb|-|latent$  & {26.62}  & {0.912}  & {0.166} \\		
			
			$z_{s}$ & {26.96}  & {0.922}  & {0.160} \\

			$z_{tv}$ & {\textbf{27.19}}  & {\textbf{0.937}}  & {\textbf{0.150}} \\
			\bottomrule[1pt]
	\end{tabular}}
\end{table}

\begin{figure}[h]
	\centering
	\includegraphics[scale=0.39]{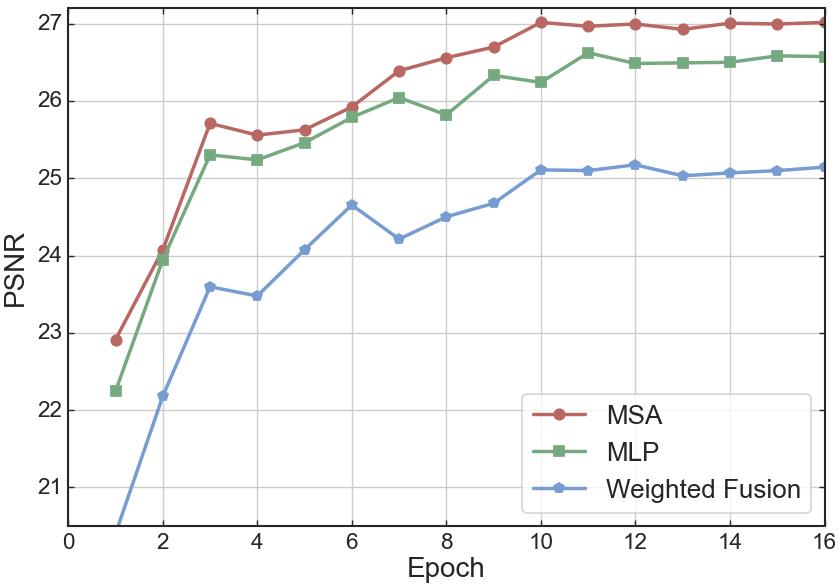}
	\caption{Comparison of different guidance approaches on the DTU dataset.}
	\label{fig:ablation_attention}
\end{figure}
\begin{figure*}[t]
	\centering
	\includegraphics[scale=0.195]{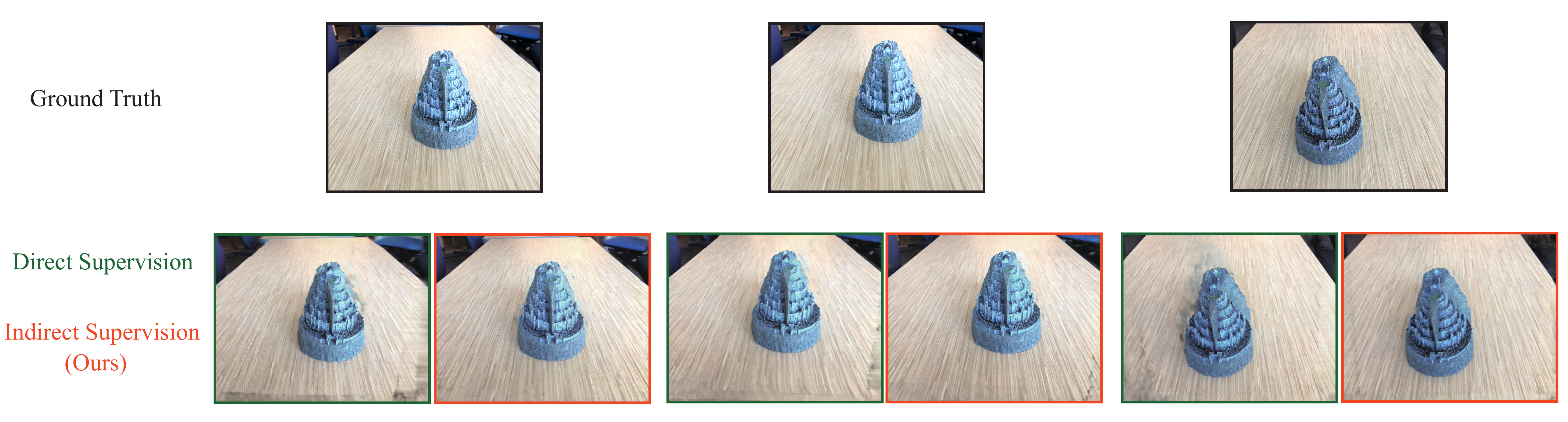}
	\caption{Qualitative comparison of two supervision approaches. Both methods are trained on the DTU dataset with 3 input views.}
	\label{fig:visualization_direct}
\end{figure*}	
\textbf{Effect of attention guidance.} %
To enable the guidance of the latent space to the reprojected features, we propose the AGM module that utilizes the MSA mechanism for this purpose. The ablation experiments are executed here to explore its effectiveness. Specifically, three different guidance strategies are devised for the latent and reprojected features: \romannumeral1) using a layer of MSA to adaptively generate weights to fuse them; \romannumeral2) inputting them directly into a layer of MLP to perform the fusion; and \romannumeral3) inputting them into a layer of MLP to generate respective weight scores, and then performing weighted fusion. The results are reported in Fig. \ref{fig:ablation_attention}, and, as can be seen, the MSA approach makes the best prediction.

\textbf{Effect of indirect supervision.} As opposed to directly supervising the renderings of NeRF using samples from the PDM-predicted distribution, our approach indirectly injects pre-trained knowledge into NeRF via a distilled latent space during training. In order to demonstrate the superiority of this indirect supervision manner, we compare it here with the direct supervision approach. Specifically, for direct supervision, the latent space $z_{tv}$ is no longer used, we use reprojected features to predict the color and density of each 3D point, and then input the rendered results directly into a pre-trained stable-diffusion to perform SDS loss. We perform experiments on the DTU dataset in the 3-input view case in sparse setting 1. We report the visualization results in Fig. \ref{fig:visualization_direct}, and as can be seen, our indirect supervision approach can make sharp renderings, while the direct supervision approach tends to make fuzzy predictions in regions such as the margins of the image and object's contours.

To analyze this, 1) the sub-optimal phenomenon of blurry predictions at the margins of the image, as discussed in Sec.~\ref{Results}, is due to erroneous reprojected features. The visualization results in Fig. \ref{fig:visualization_direct} demonstrate that our indirect manner is more advantageous than the direct manner in addressing this issue. 2) The direct method produces blurry predictions at the object's contours because the samples drawn from the PDM-generated distributions may exhibit 3D inconsistencies at the target's contours. For example, some samples may have large-sized contours while others have small-sized contours, causing the model to be confused and leading it to produce blurry, averaged predictions in these areas.

\subsection{Limitations}
Although our approach takes a promising step towards addressing the uncertainty in Gen-NeRFs, there are still some limitations that need to be addressed. For instance, there is room for improvement in terms of depth prediction and rendering quality of our generalizable model, relative to per-scene optimization methods. Additionally, while we benefit from the priors of pre-trained diffusion models, we also pay a price, such as the slow training process. This problem is worth addressing in the future and can probably be addressed by drawing inspiration from works like LCM-LoRA~\cite{LCM-LoRA}.


\section{CONCLUSIONS}

In this paper, we aim to improve the suboptimal issue caused by erroneous reprojections in current generalizable NeRFs with pre-trained generative priors. We introduce an innovative generalizable NeRF framework, called ID-NeRF, which adopts an indirect manner to inject generative priors into the generalizable NeRF paradigm. This indirect guidance strategy facilitates the transfer of knowledge from the PDM into a latent space through score-based distillation, subsequently refining the reprojected features of traditional Gen-NeRFs. Consequently, ID-NeRF not only alleviates the suboptimal problem caused by erroneous reprojected features but also effectively avoids bothersome model confusion. Our empirical results demonstrate the outstanding performance of our method across various experimental settings. In future work, we will continue to delve into how to more rationally use prior knowledge from pre-trained large models to assist generalizable novel view synthesis in sparse settings.

{\small
\bibliographystyle{ieee_fullname}
\bibliography{ref}

\begin{thebibliography}{10}\itemsep=-1pt

\bibitem{pi-gan}
Eric~R Chan, Marco Monteiro, Petr Kellnhofer, Jiajun Wu, and Gordon Wetzstein.
\newblock pi-gan: Periodic implicit generative adversarial networks for
  3d-aware image synthesis.
\newblock In {\em Proceedings of the IEEE/CVF conference on computer vision and
  pattern recognition}, pages 5799--5809, 2021.

\bibitem{mvsnerf}
Anpei Chen, Zexiang Xu, Fuqiang Zhao, Xiaoshuai Zhang, Fanbo Xiang, Jingyi Yu,
  and Hao Su.
\newblock Mvsnerf: Fast generalizable radiance field reconstruction from
  multi-view stereo.
\newblock In {\em Proceedings of the IEEE/CVF International Conference on
  Computer Vision}, pages 14124--14133, 2021.

\bibitem{virtual-try-on2}
Xinya Chen, Jiaxin Huang, Yanrui Bin, Lu Yu, and Yiyi Liao.
\newblock Veri3d: Generative vertex-based radiance fields for 3d controllable
  human image synthesis.
\newblock In {\em Proceedings of the IEEE/CVF International Conference on
  Computer Vision}, pages 8986--8997, 2023.

\bibitem{dbarf}
Yu Chen and Gim~Hee Lee.
\newblock Dbarf: Deep bundle-adjusting generalizable neural radiance fields.
\newblock In {\em Proceedings of the IEEE/CVF Conference on Computer Vision and
  Pattern Recognition}, pages 24--34, 2023.

\bibitem{MatchNeRF}
Yuedong Chen, Haofei Xu, Qianyi Wu, Chuanxia Zheng, Tat-Jen Cham, and Jianfei
  Cai.
\newblock Explicit correspondence matching for generalizable neural radiance
  fields.
\newblock {\em arXiv preprint arXiv:2304.12294}, 2023.

\bibitem{multi-face-janus2}
Lihe Ding, Shaocong Dong, Zhanpeng Huang, Zibin Wang, Yiyuan Zhang, Kaixiong
  Gong, Dan Xu, and Tianfan Xue.
\newblock Text-to-3d generation with bidirectional diffusion using both 2d and
  3d priors.
\newblock {\em arXiv preprint arXiv:2312.04963}, 2023.

\bibitem{DDPM}
Jonathan Ho, Ajay Jain, and Pieter Abbeel.
\newblock Denoising diffusion probabilistic models.
\newblock {\em Advances in neural information processing systems},
  33:6840--6851, 2020.

\bibitem{lirf}
Xin Huang, Qi Zhang, Ying Feng, Xiaoyu Li, Xuan Wang, and Qing Wang.
\newblock Local implicit ray function for generalizable radiance field
  representation.
\newblock In {\em Proceedings of the IEEE/CVF Conference on Computer Vision and
  Pattern Recognition}, pages 97--107, 2023.

\bibitem{NEO360}
Muhammad~Zubair Irshad, Sergey Zakharov, Katherine Liu, Vitor Guizilini, Thomas
  Kollar, Adrien Gaidon, Zsolt Kira, and Rares Ambrus.
\newblock Neo 360: Neural fields for sparse view synthesis of outdoor scenes.
\newblock In {\em Proceedings of the IEEE/CVF International Conference on
  Computer Vision}, pages 9187--9198, 2023.

\bibitem{DTU}
Rasmus Jensen, Anders Dahl, George Vogiatzis, Engin Tola, and Henrik Aan{\ae}s.
\newblock Large scale multi-view stereopsis evaluation.
\newblock In {\em Proceedings of the IEEE conference on computer vision and
  pattern recognition}, pages 406--413, 2014.

\bibitem{geonerf}
Mohammad~Mahdi Johari, Yann Lepoittevin, and Fran{\c{c}}ois Fleuret.
\newblock Geonerf: Generalizing nerf with geometry priors.
\newblock In {\em Proceedings of the IEEE/CVF Conference on Computer Vision and
  Pattern Recognition}, pages 18365--18375, 2022.

\bibitem{NeRF-VAE}
Adam~R Kosiorek, Heiko Strathmann, Daniel Zoran, Pol Moreno, Rosalia Schneider,
  Sona Mokr{\'a}, and Danilo~Jimenez Rezende.
\newblock Nerf-vae: A geometry aware 3d scene generative model.
\newblock In {\em International Conference on Machine Learning}, pages
  5742--5752. PMLR, 2021.

\bibitem{magic3d}
Chen-Hsuan Lin, Jun Gao, Luming Tang, Towaki Takikawa, Xiaohui Zeng, Xun Huang,
  Karsten Kreis, Sanja Fidler, Ming-Yu Liu, and Tsung-Yi Lin.
\newblock Magic3d: High-resolution text-to-3d content creation.
\newblock In {\em Proceedings of the IEEE/CVF Conference on Computer Vision and
  Pattern Recognition}, pages 300--309, 2023.

\bibitem{Lin_2023_WACV}
Kai-En Lin, Yen-Chen Lin, Wei-Sheng Lai, Tsung-Yi Lin, Yi-Chang Shih, and Ravi
  Ramamoorthi.
\newblock Vision transformer for nerf-based view synthesis from a single input
  image.
\newblock In {\em Proceedings of the IEEE/CVF Winter Conference on Applications
  of Computer Vision (WACV)}, pages 806--815, January 2023.

\bibitem{One-2-3-4}
Minghua Liu, Chao Xu, Haian Jin, Linghao Chen, Mukund Varma~T, Zexiang Xu, and
  Hao Su.
\newblock One-2-3-45: Any single image to 3d mesh in 45 seconds without
  per-shape optimization.
\newblock {\em Advances in Neural Information Processing Systems}, 36, 2024.

\bibitem{Zero-1-to-3}
Ruoshi Liu, Rundi Wu, Basile Van~Hoorick, Pavel Tokmakov, Sergey Zakharov, and
  Carl Vondrick.
\newblock Zero-1-to-3: Zero-shot one image to 3d object.
\newblock In {\em Proceedings of the IEEE/CVF International Conference on
  Computer Vision}, pages 9298--9309, 2023.

\bibitem{Neuralray}
Yuan Liu, Sida Peng, Lingjie Liu, Qianqian Wang, Peng Wang, Christian Theobalt,
  Xiaowei Zhou, and Wenping Wang.
\newblock Neural rays for occlusion-aware image-based rendering.
\newblock In {\em Proceedings of the IEEE/CVF Conference on Computer Vision and
  Pattern Recognition}, pages 7824--7833, 2022.

\bibitem{AdamW}
Ilya Loshchilov and Frank Hutter.
\newblock Decoupled weight decay regularization.
\newblock {\em arXiv preprint arXiv:1711.05101}, 2017.

\bibitem{LCM-LoRA}
Simian Luo, Yiqin Tan, Suraj Patil, Daniel Gu, Patrick von Platen,
  Apolin{\'a}rio Passos, Longbo Huang, Jian Li, and Hang Zhao.
\newblock Lcm-lora: A universal stable-diffusion acceleration module.
\newblock {\em arXiv preprint arXiv:2311.05556}, 2023.

\bibitem{occupancy}
Lars Mescheder, Michael Oechsle, Michael Niemeyer, Sebastian Nowozin, and
  Andreas Geiger.
\newblock Occupancy networks: Learning 3d reconstruction in function space.
\newblock In {\em Proceedings of the IEEE/CVF conference on computer vision and
  pattern recognition}, pages 4460--4470, 2019.

\bibitem{multi-face-janus1}
Gal Metzer, Elad Richardson, Or Patashnik, Raja Giryes, and Daniel Cohen-Or.
\newblock Latent-nerf for shape-guided generation of 3d shapes and textures.
\newblock In {\em Proceedings of the IEEE/CVF Conference on Computer Vision and
  Pattern Recognition}, pages 12663--12673, 2023.

\bibitem{latent-nerf}
Gal Metzer, Elad Richardson, Or Patashnik, Raja Giryes, and Daniel Cohen-Or.
\newblock Latent-nerf for shape-guided generation of 3d shapes and textures.
\newblock In {\em Proceedings of the IEEE/CVF Conference on Computer Vision and
  Pattern Recognition}, pages 12663--12673, 2023.

\bibitem{nerf}
Ben Mildenhall, Pratul~P Srinivasan, Matthew Tancik, Jonathan~T Barron, Ravi
  Ramamoorthi, and Ren Ng.
\newblock Nerf: Representing scenes as neural radiance fields for view
  synthesis.
\newblock {\em Communications of the ACM}, 65(1):99--106, 2021.

\bibitem{Giraffe}
Michael Niemeyer and Andreas Geiger.
\newblock Giraffe: Representing scenes as compositional generative neural
  feature fields.
\newblock In {\em Proceedings of the IEEE/CVF Conference on Computer Vision and
  Pattern Recognition}, pages 11453--11464, 2021.

\bibitem{confident_learning}
Curtis Northcutt, Lu Jiang, and Isaac Chuang.
\newblock Confident learning: Estimating uncertainty in dataset labels.
\newblock {\em Journal of Artificial Intelligence Research}, 70:1373--1411,
  2021.

\bibitem{northcutt2017learning}
Curtis~G Northcutt, Tailin Wu, and Isaac~L Chuang.
\newblock Learning with confident examples: Rank pruning for robust
  classification with noisy labels.
\newblock {\em arXiv preprint arXiv:1705.01936}, 2017.

\bibitem{deepsdf}
Jeong~Joon Park, Peter Florence, Julian Straub, Richard Newcombe, and Steven
  Lovegrove.
\newblock Deepsdf: Learning continuous signed distance functions for shape
  representation.
\newblock In {\em Proceedings of the IEEE/CVF conference on computer vision and
  pattern recognition}, pages 165--174, 2019.

\bibitem{compositional}
Ryan Po and Gordon Wetzstein.
\newblock Compositional 3d scene generation using locally conditioned
  diffusion.
\newblock {\em arXiv preprint arXiv:2303.12218}, 2023.

\bibitem{dreamfusion}
Ben Poole, Ajay Jain, Jonathan~T Barron, and Ben Mildenhall.
\newblock Dreamfusion: Text-to-3d using 2d diffusion.
\newblock {\em arXiv preprint arXiv:2209.14988}, 2022.

\bibitem{virtual-try-on1}
Yi-Ling Qiao, Alexander Gao, Yiran Xu, Yue Feng, Jia-Bin Huang, and Ming~C Lin.
\newblock Dynamic mesh-aware radiance fields.
\newblock In {\em Proceedings of the IEEE/CVF International Conference on
  Computer Vision}, pages 385--396, 2023.

\bibitem{CLIP}
Alec Radford, Jong~Wook Kim, Chris Hallacy, Aditya Ramesh, Gabriel Goh,
  Sandhini Agarwal, Girish Sastry, Amanda Askell, Pamela Mishkin, Jack Clark,
  et~al.
\newblock Learning transferable visual models from natural language
  supervision.
\newblock In {\em International conference on machine learning}, pages
  8748--8763. PMLR, 2021.

\bibitem{dreambooth3d}
Amit Raj, Srinivas Kaza, Ben Poole, Michael Niemeyer, Nataniel Ruiz, Ben
  Mildenhall, Shiran Zada, Kfir Aberman, Michael Rubinstein, Jonathan Barron,
  et~al.
\newblock Dreambooth3d: Subject-driven text-to-3d generation.
\newblock {\em arXiv preprint arXiv:2303.13508}, 2023.

\bibitem{DALLE-2}
Aditya Ramesh, Prafulla Dhariwal, Alex Nichol, Casey Chu, and Mark Chen.
\newblock Hierarchical text-conditional image generation with clip latents.
\newblock {\em arXiv preprint arXiv:2204.06125}, 1(2):3, 2022.

\bibitem{stable-diffusion}
Robin Rombach, Andreas Blattmann, Dominik Lorenz, Patrick Esser, and Bj{\"o}rn
  Ommer.
\newblock High-resolution image synthesis with latent diffusion models.
\newblock In {\em Proceedings of the IEEE/CVF conference on computer vision and
  pattern recognition}, pages 10684--10695, 2022.

\bibitem{Imagen}
Chitwan Saharia, William Chan, Saurabh Saxena, Lala Li, Jay Whang, Emily~L
  Denton, Kamyar Ghasemipour, Raphael Gontijo~Lopes, Burcu Karagol~Ayan, Tim
  Salimans, et~al.
\newblock Photorealistic text-to-image diffusion models with deep language
  understanding.
\newblock {\em Advances in Neural Information Processing Systems},
  35:36479--36494, 2022.

\bibitem{Graf}
Katja Schwarz, Yiyi Liao, Michael Niemeyer, and Andreas Geiger.
\newblock Graf: Generative radiance fields for 3d-aware image synthesis.
\newblock {\em Advances in Neural Information Processing Systems},
  33:20154--20166, 2020.

\bibitem{3DFuse}
Junyoung Seo, Wooseok Jang, Min-Seop Kwak, Jaehoon Ko, Hyeonsu Kim, Junho Kim,
  Jin-Hwa Kim, Jiyoung Lee, and Seungryong Kim.
\newblock Let 2d diffusion model know 3d-consistency for robust text-to-3d
  generation.
\newblock {\em arXiv preprint arXiv:2303.07937}, 2023.

\bibitem{OneCycle}
Leslie~N Smith and Nicholay Topin.
\newblock Super-convergence: Very fast training of neural networks using large
  learning rates.
\newblock In {\em Artificial intelligence and machine learning for multi-domain
  operations applications}, volume 11006, pages 369--386. SPIE, 2019.

\bibitem{DDIM}
Jiaming Song, Chenlin Meng, and Stefano Ermon.
\newblock Denoising diffusion implicit models.
\newblock {\em arXiv preprint arXiv:2010.02502}, 2020.

\bibitem{score-based-1}
Yang Song and Stefano Ermon.
\newblock Generative modeling by estimating gradients of the data distribution.
\newblock {\em Advances in neural information processing systems}, 32, 2019.

\bibitem{score-based-2}
Yang Song, Jascha Sohl-Dickstein, Diederik~P Kingma, Abhishek Kumar, Stefano
  Ermon, and Ben Poole.
\newblock Score-based generative modeling through stochastic differential
  equations.
\newblock 2021.

\bibitem{GPNR}
Mohammed Suhail, Carlos Esteves, Leonid Sigal, and Ameesh Makadia.
\newblock Generalizable patch-based neural rendering.
\newblock In {\em European Conference on Computer Vision}. Springer, 2022.

\bibitem{LoFTR}
Jiaming Sun, Zehong Shen, Yuang Wang, Hujun Bao, and Xiaowei Zhou.
\newblock Loftr: Detector-free local feature matching with transformers.
\newblock In {\em Proceedings of the IEEE/CVF conference on computer vision and
  pattern recognition}, pages 8922--8931, 2021.

\bibitem{GNT}
Mukund~Varma T, Peihao Wang, Xuxi Chen, Tianlong Chen, Subhashini Venugopalan,
  and Zhangyang Wang.
\newblock Is attention all that nerf needs?
\newblock In {\em The Eleventh International Conference on Learning
  Representations}, 2023.

\bibitem{stable-dreamfusion}
Jiaxiang Tang.
\newblock Stable-dreamfusion: Text-to-3d with stable-diffusion, 2022.
\newblock https://github.com/ashawkey/stable-dreamfusion.

\bibitem{avatar3}
Mukhiddin Toshpulatov, Wookey Lee, and Suan Lee.
\newblock Talking human face generation: A survey.
\newblock {\em Expert Systems with Applications}, 219:119678, 2023.

\bibitem{transformer}
Ashish Vaswani, Noam Shazeer, Niki Parmar, Jakob Uszkoreit, Llion Jones,
  Aidan~N Gomez, {\L}ukasz Kaiser, and Illia Polosukhin.
\newblock Attention is all you need.
\newblock {\em Advances in neural information processing systems}, 30, 2017.

\bibitem{sjc}
Haochen Wang, Xiaodan Du, Jiahao Li, Raymond~A Yeh, and Greg Shakhnarovich.
\newblock Score jacobian chaining: Lifting pretrained 2d diffusion models for
  3d generation.
\newblock In {\em Proceedings of the IEEE/CVF Conference on Computer Vision and
  Pattern Recognition}, pages 12619--12629, 2023.

\bibitem{ibrnet}
Qianqian Wang, Zhicheng Wang, Kyle Genova, Pratul~P Srinivasan, Howard Zhou,
  Jonathan~T Barron, Ricardo Martin-Brualla, Noah Snavely, and Thomas
  Funkhouser.
\newblock Ibrnet: Learning multi-view image-based rendering.
\newblock In {\em Proceedings of the IEEE/CVF Conference on Computer Vision and
  Pattern Recognition}, pages 4690--4699, 2021.

\bibitem{SSIM}
Zhou Wang, Alan~C Bovik, Hamid~R Sheikh, and Eero~P Simoncelli.
\newblock Image quality assessment: from error visibility to structural
  similarity.
\newblock {\em IEEE transactions on image processing}, 13(4):600--612, 2004.

\bibitem{virtual-try-on3}
Chung-Yi Weng, Brian Curless, Pratul~P Srinivasan, Jonathan~T Barron, and Ira
  Kemelmacher-Shlizerman.
\newblock Humannerf: Free-viewpoint rendering of moving people from monocular
  video.
\newblock In {\em Proceedings of the IEEE/CVF conference on computer vision and
  pattern Recognition}, pages 16210--16220, 2022.

\bibitem{su2}
Junwei Wu, Mingjie Sun, Chenru Jiang, Jiejie Liu, Jeremy Smith, and Quan Zhang.
\newblock Context-based local-global fusion network for 3d point cloud
  classification and segmentation.
\newblock {\em Expert Systems with Applications}, page 124023, 2024.

\bibitem{DiffusioNeRF}
Jamie Wynn and Daniyar Turmukhambetov.
\newblock Diffusionerf: Regularizing neural radiance fields with denoising
  diffusion models.
\newblock In {\em Proceedings of the IEEE/CVF Conference on Computer Vision and
  Pattern Recognition}, pages 4180--4189, 2023.

\bibitem{gmflow}
Haofei Xu, Jing Zhang, Jianfei Cai, Hamid Rezatofighi, and Dacheng Tao.
\newblock Gmflow: Learning optical flow via global matching.
\newblock In {\em Proceedings of the IEEE/CVF conference on computer vision and
  pattern recognition}, pages 8121--8130, 2022.

\bibitem{dream3d}
Jiale Xu, Xintao Wang, Weihao Cheng, Yan-Pei Cao, Ying Shan, Xiaohu Qie, and
  Shenghua Gao.
\newblock Dream3d: Zero-shot text-to-3d synthesis using 3d shape prior and
  text-to-image diffusion models.
\newblock In {\em Proceedings of the IEEE/CVF Conference on Computer Vision and
  Pattern Recognition}, pages 20908--20918, 2023.

\bibitem{contranerf}
Hao Yang, Lanqing Hong, Aoxue Li, Tianyang Hu, Zhenguo Li, Gim~Hee Lee, and
  Liwei Wang.
\newblock Contranerf: Generalizable neural radiance fields for
  synthetic-to-real novel view synthesis via contrastive learning.
\newblock In {\em Proceedings of the IEEE/CVF Conference on Computer Vision and
  Pattern Recognition}, pages 16508--16517, 2023.

\bibitem{FeatureNerf}
Jianglong Ye, Naiyan Wang, and Xiaolong Wang.
\newblock Featurenerf: Learning generalizable nerfs by distilling pre-trained
  vision foundation models.
\newblock {\em arXiv preprint arXiv:2303.12786}, 2023.

\bibitem{pixelnerf}
Alex Yu, Vickie Ye, Matthew Tancik, and Angjoo Kanazawa.
\newblock pixelnerf: Neural radiance fields from one or few images.
\newblock In {\em Proceedings of the IEEE/CVF Conference on Computer Vision and
  Pattern Recognition}, pages 4578--4587, 2021.

\bibitem{avatar1}
Wangbo Yu, Yanbo Fan, Yong Zhang, Xuan Wang, Fei Yin, Yunpeng Bai, Yan-Pei Cao,
  Ying Shan, Yang Wu, Zhongqian Sun, et~al.
\newblock Nofa: Nerf-based one-shot facial avatar reconstruction.
\newblock In {\em ACM SIGGRAPH 2023 Conference Proceedings}, pages 1--12, 2023.

\bibitem{su1}
Changyu Zeng, Wei Wang, Anh Nguyen, and Yutao Yue.
\newblock Self-supervised learning for point cloud data: A survey.
\newblock {\em Expert Systems with Applications}, page 121354, 2023.

\bibitem{LPIPS}
Richard Zhang, Phillip Isola, Alexei~A Efros, Eli Shechtman, and Oliver Wang.
\newblock The unreasonable effectiveness of deep features as a perceptual
  metric.
\newblock In {\em Proceedings of the IEEE conference on computer vision and
  pattern recognition}, pages 586--595, 2018.

\bibitem{nerfusion}
Xiaoshuai Zhang, Sai Bi, Kalyan Sunkavalli, Hao Su, and Zexiang Xu.
\newblock Nerfusion: Fusing radiance fields for large-scale scene
  reconstruction.
\newblock In {\em Proceedings of the IEEE/CVF Conference on Computer Vision and
  Pattern Recognition}, pages 5449--5458, 2022.

\bibitem{sparsefusion}
Zhizhuo Zhou and Shubham Tulsiani.
\newblock Sparsefusion: Distilling view-conditioned diffusion for 3d
  reconstruction.
\newblock In {\em Proceedings of the IEEE/CVF Conference on Computer Vision and
  Pattern Recognition}, pages 12588--12597, 2023.

\end{thebibliography}
}

\end{document}